\documentclass{llncs}
\usepackage{graphicx}
\usepackage{comment}
\usepackage{amsmath,amssymb} % define this before the line numbering.
\usepackage{color}
\usepackage{hyperref}
\usepackage[linesnumbered,ruled]{algorithm2e}

\usepackage[width=122mm,left=12mm,paperwidth=146mm,height=193mm,top=12mm,paperheight=217mm]{geometry}

\DeclareMathOperator*{\argmin}{arg\,min}

\newcommand{\shapeTemp}{\mathbf{S_t}}
\newcommand{\shapeQuery}{\mathbf{S_q}}
\newcommand{\pointTemp}{\mathbf{p_t}}

\newcommand{\lossChamfer}{\mathcal{L}^{\textrm{CD}}}
\newcommand{\featureEmbedding}{\mathbf{E}}
\newcommand{\correspondences}{\mathcal{C}}
\newcommand{\thetaG}{\hat{\theta}_g}

\begin{document}
\pagestyle{headings}
\mainmatter

\title{Meta Deformation Network: Meta Functionals for Shape Correspondence} % Replace with your title

% \begin{comment}
\titlerunning{Meta Deformation Network: Meta Functionals for Shape Correspondence}
\author{Daohan Lu\inst{1,4} \and
Yi Fang\inst{1,2,3}}
\authorrunning{D. Lu et al.}
\institute{
    Multimedia and Visual Computing Lab, New York University, New York, United States.\and
    Tandon School of Engineering, New York University, New York, United States.\and
    Department of Electrical and Computer Engineering, New York University, Abu Dhabi, United Arab Emirates. \and
    College of Arts and Science, New York University, New York, United States.
}
% \end{comment}
%******************
\maketitle

\begin{abstract}
We present a new technique named \textit{Meta Deformation Network} for 3D shape matching via deformation, in which a deep neural network maps a reference shape onto the parameters of a second neural network whose task is to give the correspondence between a learned template and query shape via deformation. We categorize the second neural network as a meta-function, or a function generated by another function, as its parameters are dynamically given by the first network on a per-input basis. This leads to a straightforward overall architecture and faster execution speeds, without loss in the quality of the deformation of the template. We show in our experiments that Meta Deformation Network leads to improvements on the MPI-FAUST Inter Challenge over designs that utilized a conventional decoder design that has non-dynamic parameters.

\end{abstract}

\section{Introduction}

In the pairwise 3D shape correspondence problem, a program is given two query shapes and is asked to compute all pairs of corresponding points between the two shapes. Recent works \cite{groueix20183dcoded},\cite{deprelle2019learning_elementary} yielded improved results on pairwise 3D shape correspondence by aligning a template point cloud with individual query shapes, which gives a correspondence relationship between the template and each query shape. After the correspondence between the template and each query shape was obtained, this information was used to infer the correspondence between the two query shapes. This approach usually employed an encoder-decoder design with the encoder generating a feature embedding $\mathbf{E}$ that captures the characteristics of the query shape $\mathbf{S_q}$. $\mathbf{E}$ is then fed into the decoder along with the of points on the template shape as input in order to output a set of points representing the deformation of the template into the query shape. Every point in the template is said to be in correspondence with the point on the query shape that is nearest to its deformation. Originally named in \cite{mitchell2019higherorder}, we refer to this decoder scheme as \textit{Latent Vector Concatenation} (LVC), since the input to the decoder is the concatenation of individual 3-D coordinates and the latent vector $\mathbf{E}$.

Although LVC is widely used in recent works to represent or deform 3-D shapes, \cite{park2019deepsdf},\cite{groueix20183dcoded},\cite{groueix2018atlasnet},\cite{yang2017foldingnet},\cite{deprelle2019learning_elementary}, in this paper, we investigate the possibility of an alternative decoder structure and compare it against LVC on the task of computing correspondence for human 3-D scans. Specifically, we evaluate an alternative decoder design scheme that uses only the template point cloud as input but has dynamic parameters that are predicted by a neural network from $\mathbf{E}$ and also outputs the deformed template points. We call this architecture \textit{Meta Deformation Network} because the deformation process is carried out by a neural network whose parameters are not independent but generated by another neural network. The decoder could be thought of as a second-order function that is defined or generated by another function. This formulation leads to a speedup in training and testing, and the results on the MPI-FAUST Inter correspondence challenge show that the meta decoder yields improved correspondence accuracy over a similar design \cite{deprelle2019learning_elementary} that employs an LVC decoder.

\section{Related Works}

\subsection{Dynamic Neural Networks}
Dynamic Neural Networks are a design in which one network is trained to predict the parameters of another network in test-time in order to give the second network higher adaptability. The possibility of using dynamic networks on 2-D and more recently 3-D data has been researched by several of authors. \cite{7410424_conditioned_super_res}, \cite{7299117_dynamic_weather_prediction}, \cite{bertinetto2016learning}, \cite{brab2016dynamic} applied dynamic convolutional kernels to 2-D images to input-aware single-image super-resolution, short-range weather prediction, one-shot character classification and tracking, few-shot video frame prediction, respectively. These tasks demanded highly adaptive behaviors from the neural networks, and they show improvements when using dynamic neural networks over conventional methods that used static-parameter convolutional kernels. \cite{littwin2019deep} was one of the first works to apply the idea of dynamically generated parameters to the 3-D domain, where it achieved state-of-the-art results on single-image 3D reconstruction on ShapeNet-core \cite{chang2015shapenet} compared to conventional methods in \cite{zeng2018inferring}, \cite{fan2016point}, \cite{richter2018matryoshka}.
We are motivated by these works to further investigate the potential of dynamic neural networks. We build on previous studies by examining the effectiveness of dynamic neural networks for 3-D shape correspondence in which both the input shapes and output correspondence are of a 3-dimensional nature. 

\subsection{3-D Shape Correspondence}
The registration of and correspondence among 3-D structures with non-rigid geometrical variations is a problem that has been extensively studied and one in which machine-learned based methods have had great success. MPI-FAUST \cite{Bogo:CVPR:2014} is a widely used benchmark for finding correspondence between pairs of 3-D scans of humans with realistic noises and artifacts as a result of scanning real-world humans. There are various approaches to the task. \cite{littwin2019deep} used a Siamese structure to learn compact descriptors of input shapes and using functional maps \cite{functional_maps} to solve for inter-shape correspondences. \cite{Zuffi_2015_CVPR} fitting query meshes to a pre-designed part-based model ("stitched puppet") and aligning the models for the two query scans to get correspondence. \cite{groueix20183dcoded} took a simpler approach, by choosing a set of points to be a template and getting inter-shape correspondences by computing the query shapes' individual correspondences to the common template, effectively factoring the problem of computing correspondence between two unknown shapes into two problems each involving the correspondence between a known template and an unknown shape. This method proved to give the best correspondence accuracy on FAUST and is the inspiration for the training procedure of our method, the Meta Deformation Network. \cite{groueix20183dcoded} obtains correspondence between a template and a query shape by using a multi-layer perceptron $g$ to deform the points on templates into the shape of the query and obtaining correspondence by Euclidean proximity between the deformed points of the template and actual points on the query shape. However, $g$ holds fixed parameters and takes a latent vector concatenated with template points as input. By contrast, our Meta Deformation Network has a decoder $g$ that holds dynamic parameters, which adds more flexibility and adaptability of the deformation with respect to different query shapes.

\section{Network Design}
\begin{figure}
    \centering
    \includegraphics[width=12cm]{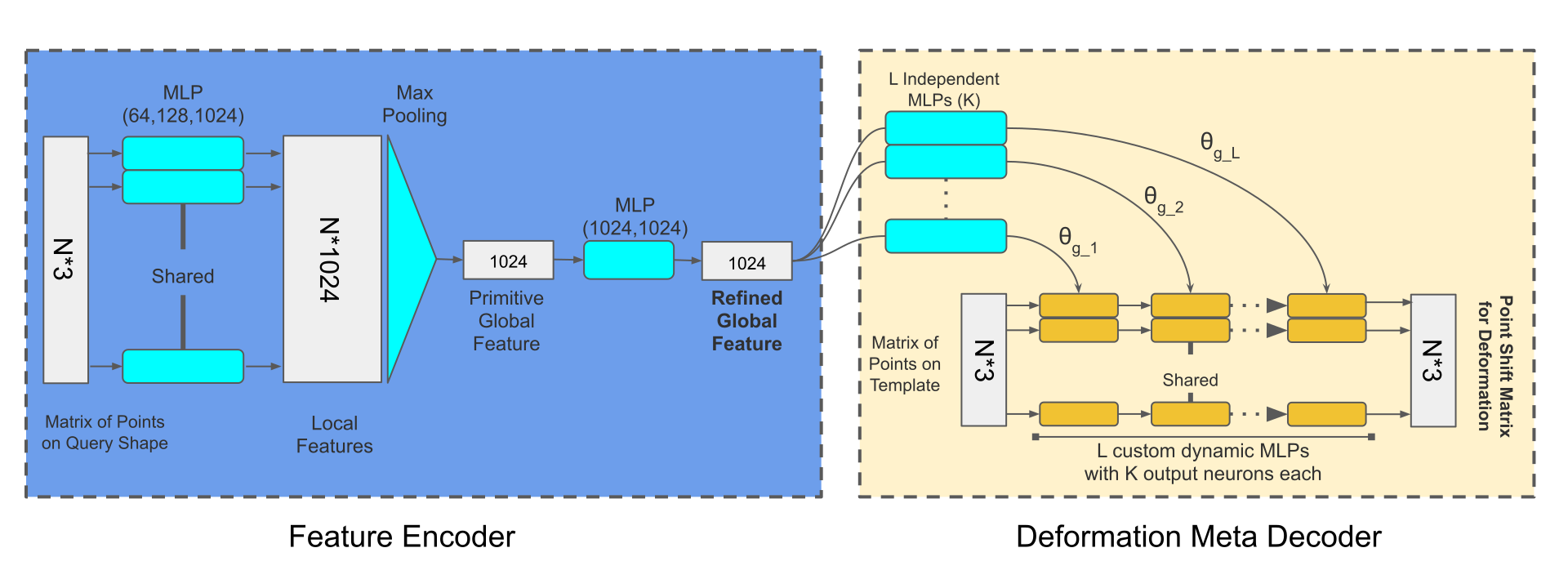}
    \caption{Architecture Overview of the Meta Deformation Network. Blue rounded rectangles represent multi-layer perceptrons (MLPs) with numbers in parentheses specifying the number of neurons for all layers, yellow rounded rectangles represent dynamic MLPs, and white rectangles indicate feature sizes at different stages in the pipeline}
    \label{fig:main_pipeline}
\end{figure}
\subsection{Architecture Overview}
\label{sec:architecture_overview}
A graphical description of the Meta Deformation Network is shown in fig. \ref{fig:main_pipeline}.
We used a simplified version of the PointNet \cite{qi2016pointnet} encoder following \cite{groueix20183dcoded}, which we will denote as $f$, to output a feature vector $\mathbf{E}$ of size 1024 from a set of points from the query shape $\mathbf{S_q}$. L independent MLPs, each denoted $h_i$, then $\mathbf{E}$ takes as input outputs $\hat{\theta_g}_i$, the predicted optimal parameters for layer $i \in \{1,...,L\}$ of the meta decoder $g$. If layer $i$ of $g$ has $K_{in}$ input neurons and $K_{out}$ output neurons, then $\hat{\theta_g}_i$ is a vector of size $(K_{in}*K_{out} + K_{out} + K_{out})$ since we define the forward-pass for each layer of g as
\begin{equation}
g_i: x_{i+1} = (W_ix_{i})*s_i + b_i
\label{eq:meta_decoder}
\end{equation}
where $*$ denotes element-wise multiplication, and $x_{out}$ is the pre-activation output. Note that we place a scaling term $s$ in addition to the typical feed-forward calculation of an MLP with the intuition that it facilitates the learning of $h$ (reasoning behind this intuition is given in \ref{sec:decoder_details}). 
\newline

With $\hat{\theta}_i$ computed, the meta decoder takes as input a point $\mathbf{p_t} \in \mathbf{S_t}$ outputs a 3-D residual vector $\Delta(\mathbf{p_t})$ that will shift $\mathbf{p_t}$ to the location of the corresponding point on the query shape $\mathbf{S_q}$. Note that the input layer to $g$ is only a vector of 3 elements, which leads to a major speed up over a decoder that uses the LVC input, which take in a vector of size $1024+3$ if using an $\featureEmbedding$. The difference in time is greater as the resolution of the template increases because the deformation computation is repeated $N$ times to get the respective $\Delta(\mathbf{p_t})$ for all $\pointTemp \in \shapeTemp$.

In all experiments, we pick $g$'s structure $L = 6$ and $K_{out}=64$ for every layer of $g$ except the last, which has $K_{out}=3$ since it outputs a 3 dimensional translation vector $\Delta{\mathbf{p_t}}$ for each input point $\mathbf{p_t}$.

\subsection{Encoder}
We used a simplified variant of the PointNet feature encoder \cite{qi2016pointnet}. The encoder, denoted $f$, applies a shared 3-layered MLP of output sizes $(64, 128, 1024)$ to each point $\mathbf{p_q} \in \mathbf{S_q}$ to obtain a local feature of size $(N\times1024)$ where $N$ is the number of points chosen from the template and is equal to $6890$ in all experiments. It then applies max-pooling along the first dimension to obtain a primitive global feature vector of size $(1024)$, which is refined by a 2-layered MLP of output sizes $(1024, 1024)$ to become the final global feature embedding $\mathbf{E}$. Formally,
\begin{equation}
    \begin{aligned}
    \mathbf{E} = f_{\theta_f}(\mathbf{S_q})
    \end{aligned}
    \label{eq:feature_encoder}
\end{equation}
We pick $N = 6980$ for all of our experiments.

\subsection{Parameter Predictor for Meta Decoder}
Having defined operations performed by each layer of $g$, we define a set of 1-layered MLPs (or more accurately, SLPs), which we will denote as $h_i$, that maps the global feature embedding $\mathbf{E}$ to $\hat{\theta}_i$. Formally,
\begin{equation}
    \begin{aligned}
  \hat{\theta}_{g_i} = h_{i, \theta_{h_i}}(\mathbf{E}) = (W_i; s_i; b_i)\\
  \hat{\theta}_g = \{\hat{\theta}_{g_i}: i \in \{1,...,L\}\}
  \label{eq:parameter_mapper}
  \end{aligned}
\end{equation}
So $h_i$ has an input size of $1024$ and an output size that matches the number of parameters needed for the $i^{th}$ layer of $g$ (see the calculation above eq. \ref{eq:meta_decoder} in \ref{sec:architecture_overview}).

\begin{figure}
    \centering
    \includegraphics[width=2.5cm]{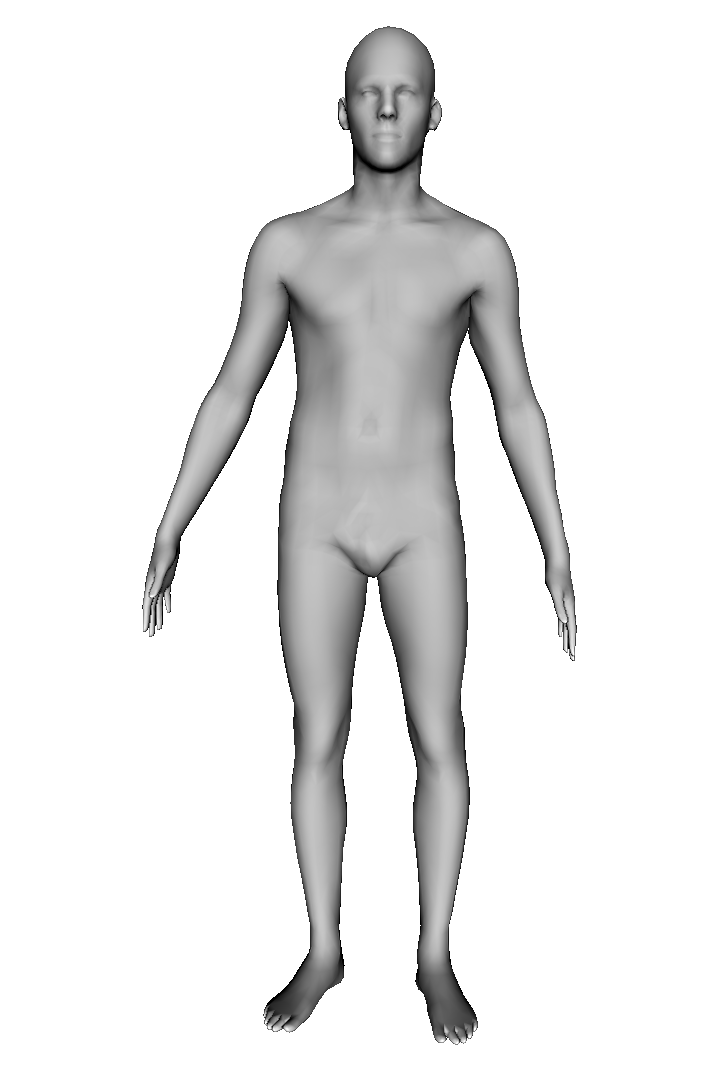}
    \includegraphics[width=1.9cm]{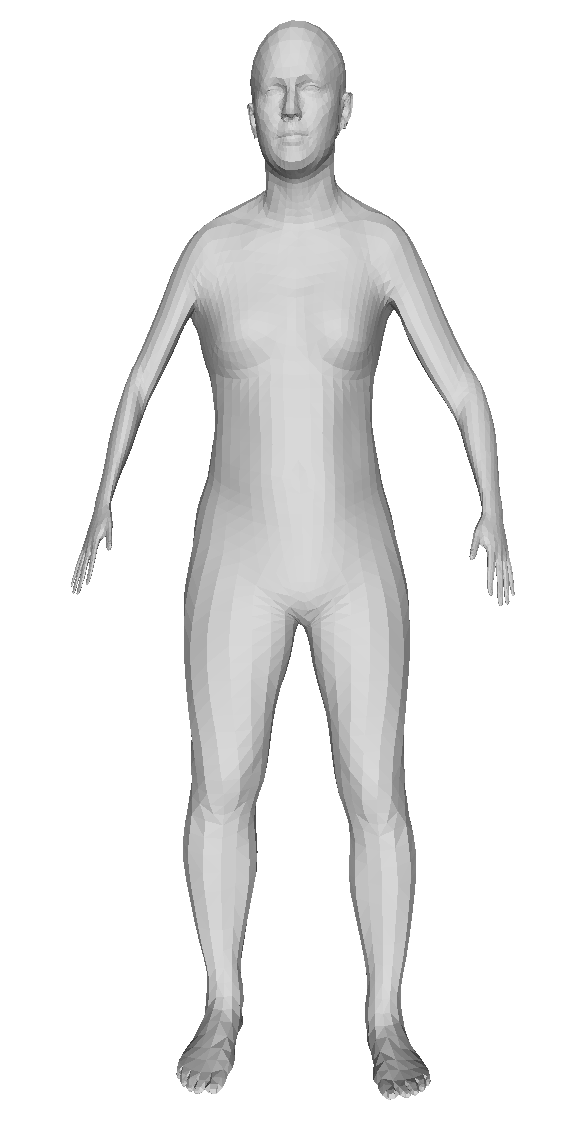}
    \caption{Learnable Template. From left to right: initialization, template learned after 40 epochs of training}
    \label{fig:template_init}
\end{figure}

\subsection{Learnable Template}
Following \cite{deprelle2019learning_elementary}, instead of choosing a template prior to training and making it permanently fixed, we initialize our template with a subsampled scan (so that $N = 6890$) from the MPI-FAUST training dataset and learns a translation vector $v_i$ for each original point on the template $\mathbf{q_{t_i}}$. So $\mathbf{p_{t_i}} = \mathbf{q_{t_i}} + v_i, i \in \{1, ..., N\}$. This allows the network to learn an template that results in the best deformation correspondence during training. The chosen template initialization is depicted in fig \ref{fig:template_init}. In test-time, we hold the template fixed by fixing the learned translation vectors $\{v_i\}$. Note that in test-time, if a high-resolution template is desired, we will no longer be able to use the learned translation vectors $\{v_i\}_{i=1}^{N}$ because each $v_i$ is learned specifically for $\pointTemp_i$. We will not have a dense set of translation vectors for every point in $\{p'_{t_j}\}_{j=1}^{N'}$ if we wish to use a high-resolution version (i.e. $N' > N$) of the initial template in test-time.

\subsection{Meta Decoder}
\label{sec:decoder_details}
We formulate each layer of $g$ as a customized version of a fully-connected layer (from eq. \ref{eq:meta_decoder}),
\begin{equation}
  g_i: x_{i+1} = (W_ix_{i})*s_i + b_i
  \tag{\ref{eq:meta_decoder}}
\end{equation}
% Although multiplying $Wx_{in}$ by $s$ is equivalent to scaling each $i^{th}$ row of $W$ by $s_i$,  the reason for the inclusion of $s$ is that it facilitates faster learning of the $h$ that maps $\mathbf{E}$ to $\hat{\theta}_g$ by allowing it to predict a single number $s_i$ instead of changing an entire row of $W$ in order to scale the $i^{th}$ element of $Wx_{in}$ by different factors, which may often be required when representing shapes with varying surface normals.
The full network $g$ outputs a residual vector for every input point $\mathbf{p_t} \in \mathbf{S_t}$ which will take it to the location of the corresponding point $\mathbf{p_q} \in \mathbf{S_q}$. Formally,
\begin{equation}
    \begin{aligned}
    \Delta(\mathbf{p_t}) = g_{\hat{\theta}_g}(\mathbf{p_t}) \\
    \hat{\mathbf{p}}_q = \mathbf{p_t} + \Delta(\mathbf{p_t})
    \end{aligned}
    \label{eq:translative_deformation}
\end{equation}
Optimizing correspondence among query shapes is equivalent to minimizing the distance between $\hat{\mathbf{p}}_q$ and $\mathbf{p}_q$ for every query shape. Thus, the formulations above leads to a straightforward supervised training loss with just a single term when training for correspondence.

\subsection{Training and Losses}
Since $\hat{\theta}_g$ is predicted, we only need to update the parameters of $f$ and $h$ in train-time. Let $\mathbb{S}_q$ be the set of all query shapes in the training dataset, this gives the following supervised optimization problem, which is the same equation used in \cite{groueix20183dcoded}:
\begin{equation}
    \begin{aligned}
    \min_{\theta_f, \theta_h} \sum_{\mathbf{S_q} \in \mathbb{S}_q}
    \sum_{i=1}^{N}
    ||g_{\hat{\theta}_g}(\mathbf{p_{t,i}}) - \mathbf{p_{q,i}}||^2 \\
    \text{where } \hat{\theta}_g = h_{\theta_h}(\mathbf{E}), \mathbf{E} = f_{\theta_f}(\mathbf{S_q})
  \label{eq:optimization}
  \end{aligned}
\end{equation}
Note that in the supervised case we assume having knowledge of the ground-truth location of the point $\mathbf{p_{q,i}} \in \mathbf{S_q}$ in correspondence with $\mathbf{p_{t,i}} \in \mathbf{S_t}$ for all $i \in {1, ..., N}$. However, we do not need any explicit information on the correspondence among the query shapes themselves as it is implicitly given by the correspondence with a common template.

\begin{figure}
    \centering
    \includegraphics[width=2.6cm]{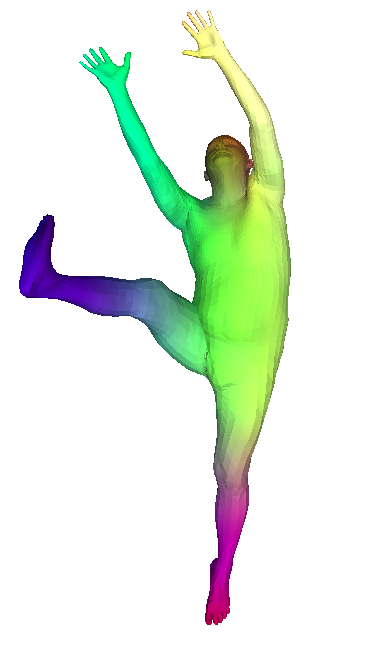}
    \includegraphics[width=3cm]{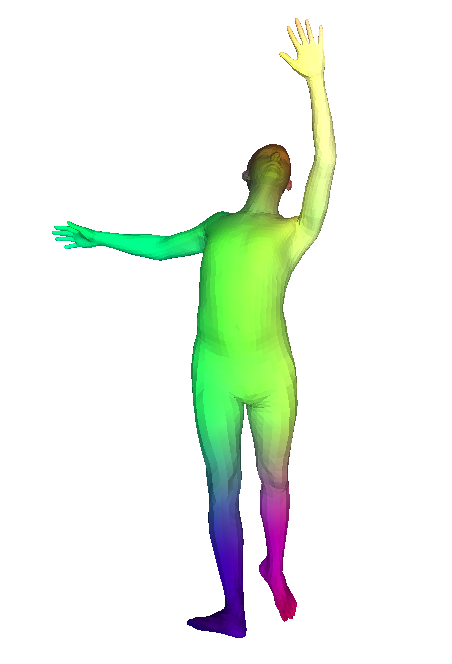}
    \includegraphics[width=3cm]{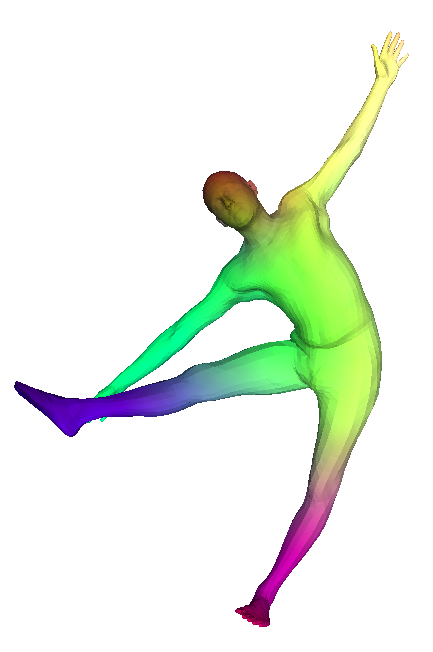}
    \caption{Query Shapes from the training dataset by \cite{groueix20183dcoded}}
    \label{fig:training_dataset}
\end{figure}

We adopt augmented SURREAL dataset by created \index{Groueix, Thibault}T. Groueix et al. \cite{groueix20183dcoded} using SMPL \cite{SMPL:2015} and SURREAL \cite{varol17_surreal}, containing 229,984 meshes of synthetic humans in various poses for training and 200 for validation. Samples of the training data are shown in fig. \ref{fig:training_dataset}. We train the model using ADAM with at a learning rate of $2*10^{-5}$ for $25$ epochs and then $2*10^{-5}$ for $15$ epochs. This takes around $14$ hours on a machine with a GTX 1080 Ti and a 6-core Intel CPU. 

\section{Inferencing Correspondences}

\subsection{Optimizing Query Shape Rotation}
As in \cite{groueix20183dcoded}, we use a procedure during testing to improve the deformation before the optimization of the latent vector $\featureEmbedding$. Before calculating the correspondence, we rotate each query shape $\shapeQuery$ by selecting different pitch and yaw angles so that the deformation of the template given the rotated $\shapeQuery$ has the smallest Chamfer distance to the rotated $\shapeQuery$. (When returning the correspondence output, we apply the inverse rotation matrices to recover the pre-rotation predicted locations).

\begin{align}
    \shapeQuery \leftarrow \argmin_{\alpha, \beta} \lossChamfer(h((R_{z,\beta}R_{y,\alpha})\shapeQuery), (R_{z,\beta}R_{y,\alpha})\shapeQuery)
\end{align}
where $R_{z,\beta}$ stands for the counterclockwise rotation matrix around the $z$ axis by angle $\beta$ (yaw), $R_{y,\alpha}$ stands for the counterclockwise rotation matrix around the $y$ axis (pitch) by angle $\alpha$, and $(R_{z,\beta}R_{y,\alpha})\shapeQuery$ denotes the set of points after applying the pitch and yaw rotation matrices to every point in $\shapeQuery$. 
In practice, we try every combination of $(\alpha, \beta)$ where $alpha \in [-\pi / 2, \pi / 2]$, $\beta \in [-\pi / 4, \pi / 4]$. The intervals are discrete with strides $\pi / 100$ and $\pi / 50$ respectively for $\alpha$ and $\beta$. This leads to a total of $100*25$ tries during which we remember the pair $(\alpha^*, \beta^*)$ that gives the best deformation and replace $\shapeQuery$ with $(R_{z,\beta^*}R_{y,\alpha^*})\shapeQuery$.

\subsection{Latent Vector Optimization}
As \ref{sec:qualitative_1} illustrates, using $f(\mathbf{S_q})$ directly as the feature embedding $\mathbf{E}$ produces suboptimal deformations. To mitigate this problem, during testing only, we use $f(\mathbf{S_q})$ as an initialization for $\mathbf{E}$ and optimizes over $\mathbf{E}$ for 3000 iterations using ADAM at a learning rate of $5*10^{-5}$ to implicitly find the $\hat{\theta}_g$ that minimizes the Chamfer distance between the deformation $\{g_{\hat{\theta}_g}(\mathbf{p_t}) : \mathbf{p_t} \in \mathbf{S_t} \}$ and the query shape $\{\mathbf{p_q} : \mathbf{p_q} \in \mathbf{S_q} \}$. We do not perform this step in training as we reduces training speed and we want to train the encoder to predict feature embeddings directly.
The Chamfer loss given $\featureEmbedding$ and $\shapeQuery$ is defined as:
\begin{equation}
\begin{aligned}
    \lossChamfer(\featureEmbedding, \shapeQuery) = \sum_{\mathbf{p_t} \in \mathbf{S_t}} \min_{\mathbf{p_q} \in \mathbf{S_q}}||\mathbf{p_q} - g_{h(\mathbf{E})}(\mathbf{p_t})||^2 +
    \sum_{\mathbf{p_q} \in \mathbf{S_q}} \min_{\mathbf{p_t} \in \mathbf{S_t}} ||\mathbf{p_q} - g_{h(\mathbf{E})}(\mathbf{p_t})||^2\\
    % \correspondences &= \{( \mathbf{p_t}, g_{h(\mathbf{E_{opt}})}(\mathbf{p_t}) ): \mathbf{p_t} \in \mathbf{S_t}\}
\end{aligned}
\label{eq:loss_chamfer}
\end{equation}

\subsection{Finding Pairwise Correspondences}
We use the algorithm presented in \cite{groueix20183dcoded} to optimize over $\mathbf{E}$ and find the correspondence between a pair of 3-D shapes, shown here in algorithm \ref{alg:inference}. Note that although we only compute correspondence between two shapes for the FAUST-Inter challenge, the same algorithm can easily be extended to predict correspondence across multiple shapes with an execution time linear to the number of query shapes by matching each shape to the common template.
% Given $S_q$ in test time,
% \begin{equation}
% \begin{aligned}
%     \mathbf{E_{init}} &= f(\mathbf{S_q})\\
%     \mathbf{E_{opt}} &= \argmin_{\mathbf{E}} \sum_{\mathbf{p_t} \in \mathbf{S_t}} \min_{\mathbf{p_q} \in \mathbf{S_q}}||\mathbf{p_q} - g_{h(\mathbf{E})}(\mathbf{p_t})||^2 +
%     \sum_{\mathbf{p_q} \in \mathbf{S_q}} \min_{\mathbf{p_t} \in \mathbf{S_t}} ||\mathbf{p_q} - g_{h(\mathbf{E})}(\mathbf{p_t})||^2\\
%     \correspondences &= \{( \mathbf{p_t}, g_{h(\mathbf{E_{opt}})}(\mathbf{p_t}) ): \mathbf{p_t} \in \mathbf{S_t}\}
% \end{aligned}
% 
% \end{equation}
% Where $\correspondences$ is the set of all corresponding points between $\mathbf{S_t}$ and $\mathbf{S_q}$ .

\begin{algorithm}[t]
\caption{Algorithm for finding 3D shape correspondences \cite{groueix20183dcoded}}
\label{alg:inference}
\SetKwInOut{Input}{Input}
\SetKwInOut{Output}{Output}
\Input{Query shape 1 $\mathbf{S_{q1}}$ and query shape 2 $\mathbf{S_{q2}}$}
\Output{Set of 3D point correspondences $\correspondences$}
\#Regression steps over latent code to find best deformation into $\mathbf{S_{q1}}$ and $\mathbf{S_{q2}}$  \\
$\mathbf{E_{1}}\leftarrow \argmin_{\mathbf{E}} \lossChamfer(\featureEmbedding, \mathbf{S_{q1}})$  \#$\featureEmbedding$ initialized with $f(\mathbf{S_{q1}})$ \\
$\mathbf{E_{2}}\leftarrow \argmin_{\mathbf{E}} \lossChamfer(\featureEmbedding, \mathbf{S_{q2}})$ \#$\featureEmbedding$ initialized with $f(\mathbf{S_{q1}})$ \\
$(\thetaG)_1 \leftarrow h(\mathbf{E_{1}})$ \\
$(\thetaG)_2 \leftarrow h(\mathbf{E_{2}})$ \\
$\correspondences\leftarrow\varnothing$ \\
\# Matching of $\mathbf{p_{q1}} \in \mathbf{S_{q1}}$ to  $\mathbf{p_{q2}} \in \mathbf{S_{q2}}$ \\
\ForEach{$\mathbf{p_q1} \in \mathbf{S_q1}$}
{
    $\hat{\mathbf{p_t}}\leftarrow \argmin_{\mathbf{p_t}\in\mathbf{S_t}} ||g_{(\thetaG)_1}(\mathbf{p_t}) - \mathbf{p_{q1}} ||^2$ \\
    $\hat{\mathbf{p_{q2}}}\leftarrow \argmin_{\mathbf{p_{q2}}\in\mathbf{S_{q2}}} ||g_{(\thetaG)_2}(\hat{\mathbf{p_t}}) - \mathbf{p_{q2}} ||^2$ \\
    $\correspondences\leftarrow\correspondences\cup\{(\mathbf{p_{q1}}, \hat{\mathbf{p_{q2}}})\}$
}
return $\correspondences$
\end{algorithm}

\section{Analysis on the Characteristics of the "Meta" Decoder}
\subsection{Defining the Meta Decoder}
In a Latent Vector Concatenation decoder construction, a fixed-parameter decoder $g$ takes as input the coordinates of a template point $\mathbf{p_t} \in \mathbf{S_t}$ concatenated with the feature embedding $\mathbf{E_q}$ representing the query shape $\mathbf{S_q}$ and outputs the predicted coordinates $\mathbf{\hat{p}_q}$ of the corresponding point in the query shape. Since we predict pairwise correspondence based on the query shapes' individual correspondences to the common template, the quality of the template deformation was crucial to the correspondence accuracy. In LVC, because the decoder has fixed parameters and needs to accurately deform all points on the template into corresponding points on various query shapes when given only $\mathbf{p_t}$ and $\mathbf{E_q}$ as input, its structure is often complex in terms of the number of trainable parameters, which could leads to problems like lengthy training times and the overfitting. Denoting the feature encoder as $f$ and the decoder as $g$, this leads to the formulation:
\begin{align}
\mathbf{E_q} &= f_{\theta_f}(\mathbf{S_q}) \\
\mathbf{\hat{p}_q} &= g_{\theta_g}(\mathbf{p_t}; \mathbf{E_q})
\end{align}
Where $\theta_f$ and $\theta_g$ are jointly optimized during training.

Our approach simplifies the structure of the decoder by formulating it as a concise dynamic multi-layer perceptron, that is, a MLP whose parameters are dynamically and uniquely determined by another neural network for each input. Denoting the encoder f and the dynamic decoder g, we have:
\begin{align}
\hat{\theta}_g &= f_{\theta_f}(\mathbf{S_q})\\
\mathbf{\hat{p}_q} &= g_{\hat{\theta}_g}(\mathbf{p_t}) + \mathbf{p_t} 
\end{align}
Because $\hat{\theta}_g$ is determined by f, we only need to optimize $\theta_f$ during training. Note that the decoder computes a location residual instead of the coordinates of the actual deformed point $\mathbf{\hat{p}_q}$ that corresponds to $\pointTemp$. We found that predicting the residual was an easier task to learn for the network. 

\subsection{Mathematical differences between LVC and Meta Decoder}
Although they perform different tasks, the neural networks in \cite{groueix20183dcoded}, \cite{qi2016pointnet}, and \cite{park2019deepsdf} have all employed $g$ as an LVC decoder that is an MLP with non-dynamic parameters. E. Mitchell pointed out in \cite{mitchell2019higherorder} that this formulation is equivalent to having a MLP that has fixed input and weights but a variable bias. Consider a case in which $g$ needs to deform a template into query shapes, then the LVC deformation decoder $g$ would take in of $\mathbf{p_t} \in \mathbf{S_t}$ concatenated with a feature embedding representing the query shape $\mathbf{E} = f_{\theta_f}(\mathbf{S_q})$. Such is the case in \cite{groueix20183dcoded}. Let vector $x$ be the concatenation of $\mathbf{p_t}$ and $\mathbf{E}$, that is, $x = (\mathbf{p_t};\mathbf{E})$, then the first layer of the LVC MLP decoder will perform the following operation:
\begin{equation}
    \begin{aligned}
    a = Wx + b
    \end{aligned}
\end{equation}
\newline
where $a$ is the pre-activation output, $W$ is the fixed weights matrix, and $b$ is the fixed bias. We can rewrite $x$ as the sum of two vectors $y$ and $z$ where $y = (\mathbf{p_{t,x}}, \mathbf{p_{t,y}}, \mathbf{p_{t,z}}, 0, 0, ...)$ and $z = (0, 0, 0, \mathbf{E})$
\begin{equation}
    \begin{aligned}
    a = Wx + b = W(y+z) + b = Wy + Wz + b
    \end{aligned}
\end{equation}
\newline
Note that $W,b,x$ are fixed with regards to $\mathbf{S_q}$ and only z changes with different $\mathbf{S_q}$, so the only term that varies with the query shape is $Wz$. Thus, $Wz$ can be seen as a dynamic bias predicted from query shapes for the first layer, which has fixed weights and input (points on the template were fixed in test-time). Using the concatenation of $\mathbf{p_{t}}$ and $\mathbf{E}$ as input is therefore equivalent to having an MLP that has fixed input, weights, and biases, except that a variable bias term $Wz$ is added to layer 1's pre-activation output that depends on $\mathbf{S_q}$. The low adaptability of the LVC decoder prevents it from producing high-quality deformations with a concise structure. 

In the case of the meta decoder, we set all parameters of $g$ to be dynamic. With this approach, the input $x$ is simply a 3-D vector $(\mathbf{p_{t,x}}, \mathbf{p_{t,y}}, \mathbf{p_{t,z}})$, and the first layer of the decoder MLP performs the operation
\begin{equation}
    \begin{aligned}
    a = (Wx)*s + b \\
    % (W; s; b) = h_1(f(\mathbf{S_q}))
    \end{aligned}
\end{equation}
where $W, s, b$ are given by $h_1(f(\mathbf{S_q}))$. The process is similar for subsequent layers. As a result, all parameters of $g$ are dynamic a function of the query shape $\mathbf{S_q}$. Defining $g$'s parameters to be a function of $\mathbf{S_q}$ offers the decoder more flexibility in computing the optimal translation that deforms the template into diverse query shapes. We also include an element-wise scaling factor $s$ in the formulation of $g$ to facilitate the learning of $h$, which maps $\mathbf{E}$ to $\hat{\theta}_g$. To see this, suppose that a particular output neuron of $g$ needs to be changed in order to compute the optimal deformation. Multiplying $s_1$ by $\lambda$ has the same effect on the output as scaling all elements in the first row of $W$ by $\lambda$, but is an easier change to predict for $h$ since it requires predicting a single number instead of multiple numbers to capture this change. Overall, the meta decoder has fully adaptive parameters to different query shapes. By contrast, the LVC decoder has only a variable bias in the output of the first layer and fixed parameters everywhere else. The enhanced adaptability of $g$ with respect to query shapes allows it to produce better deformations of the template with fewer parameters and at higher speeds.

\section{Experiment Results}

We test the Meta Deformation Network on the MPI-FAUST Inter and Intra Subject Challenges. The "inter" challenge contains 40 pairs of 3D scans of real people with each pair consisting of two different people at different poses. In test-time, to get the optimal deformation, we use $f(\mathbf{S_q})$ as an initialization for the feature embedding $\mathbf{E}$, and optimizes over $\mathbf{E}$ for 3000 iterations using ADAM at a learning rate of $5*10^{-5}$ to implicitly find the $\hat{\theta}_g$ that minimizes the Chamfer distance between the deformation $\{g_{\hat{\theta}_g}(\mathbf{p_t}) : \mathbf{p_t} \in \mathbf{S_t} \}$ and the query shape $\{\mathbf{p_q} : \mathbf{p_q} \in \mathbf{S_q} \}$.

\subsection{Unoptimized Deformations}
\label{sec:qualitative_1}
\begin{figure}[]
    \centering
    \includegraphics[height=3.2cm]{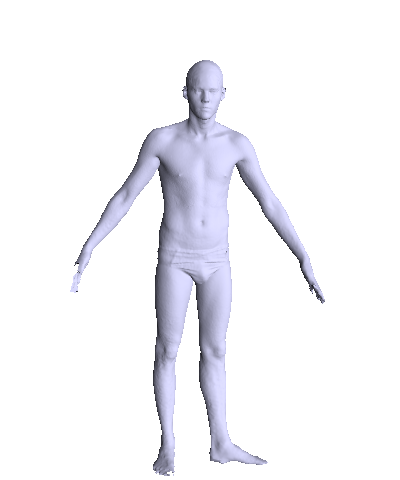}
    \includegraphics[height=3.0cm]{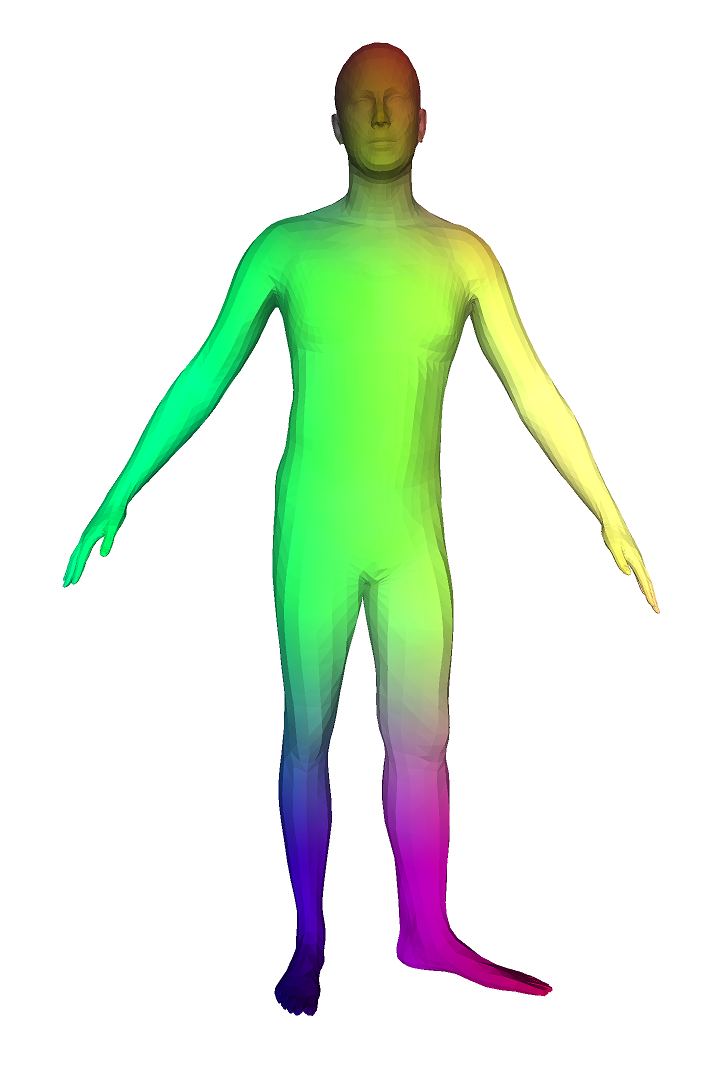}
    \includegraphics[height=3.0cm]{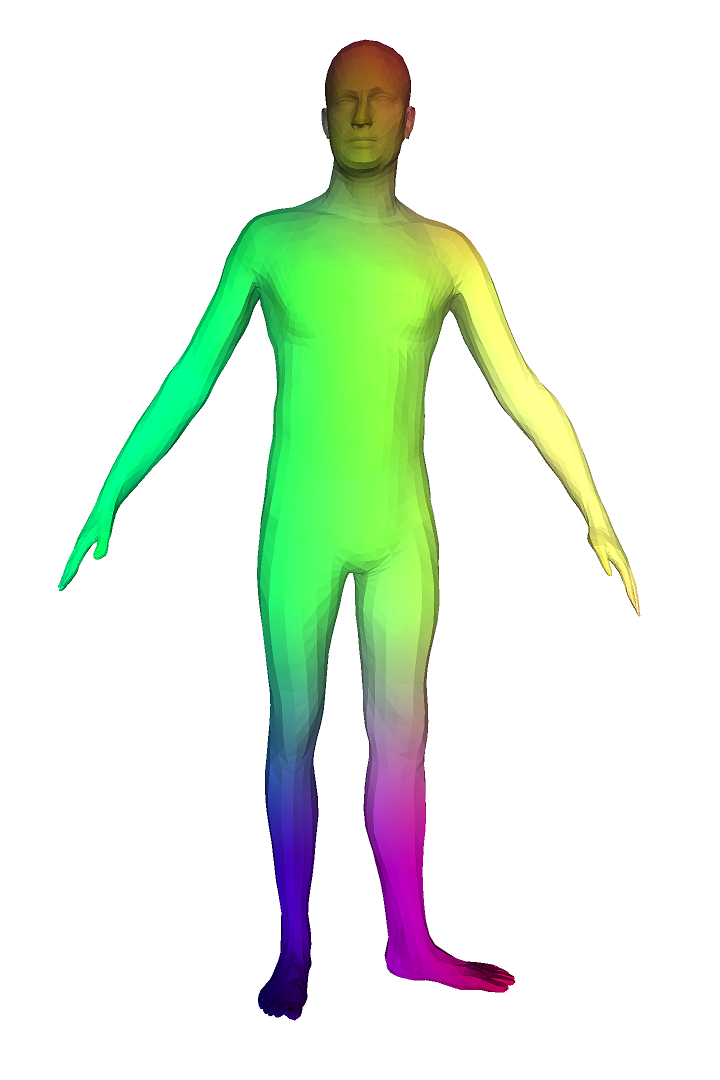}
    
    \includegraphics[height=3.2cm]{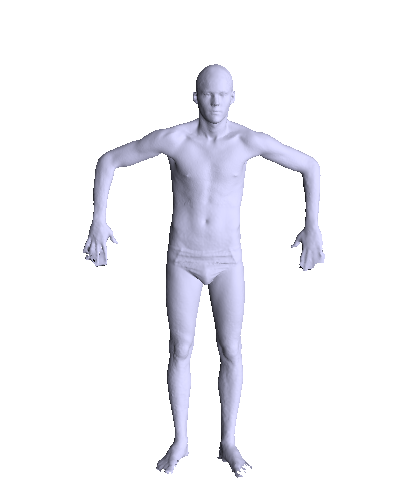}
    \includegraphics[height=3.0cm]{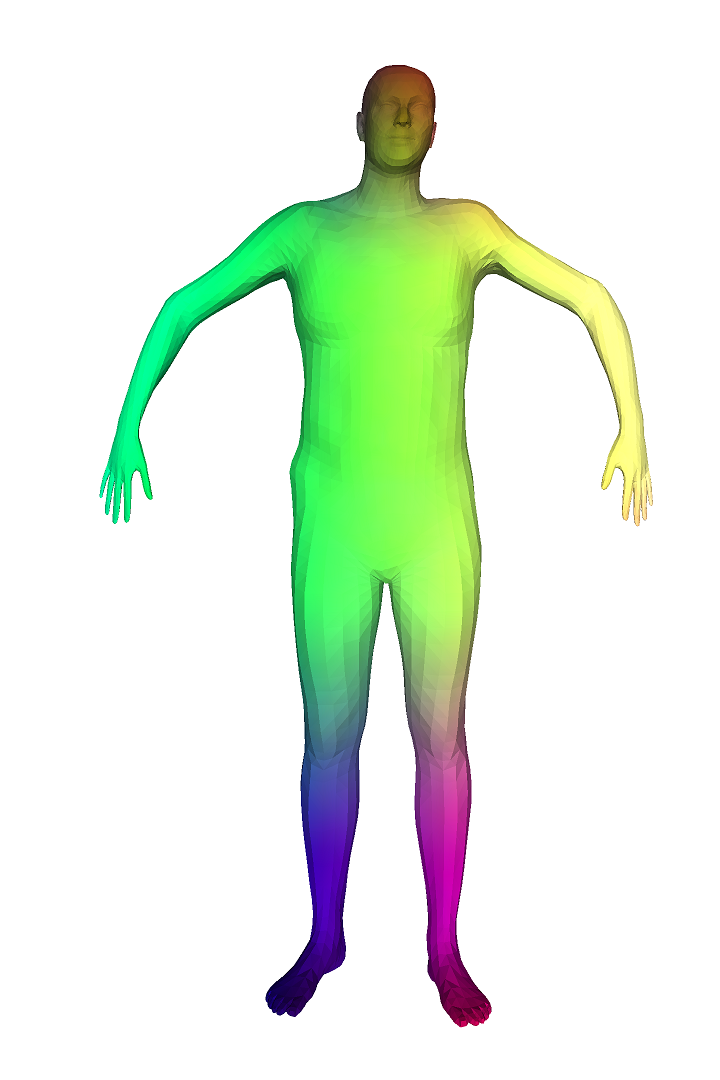}
    \includegraphics[height=3.0cm]{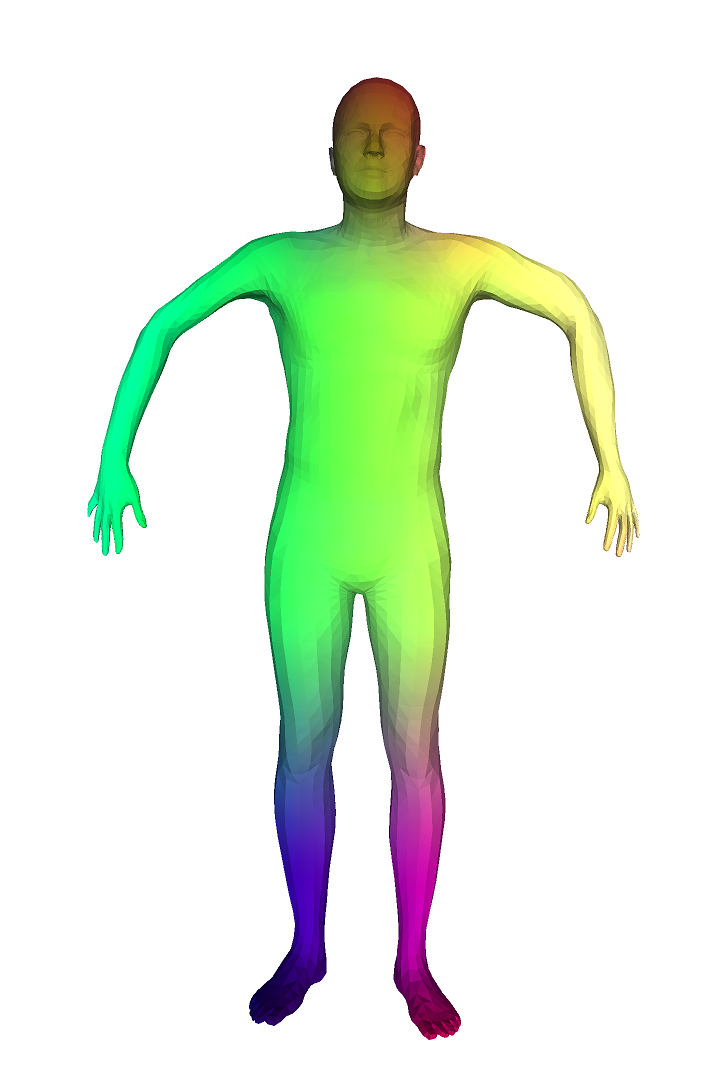}
    
    \includegraphics[height=3.2cm]{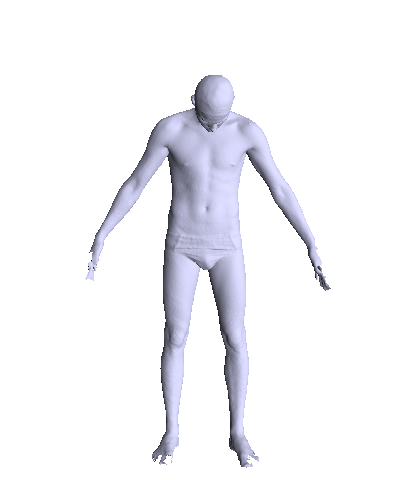}
    \includegraphics[height=3.0cm]{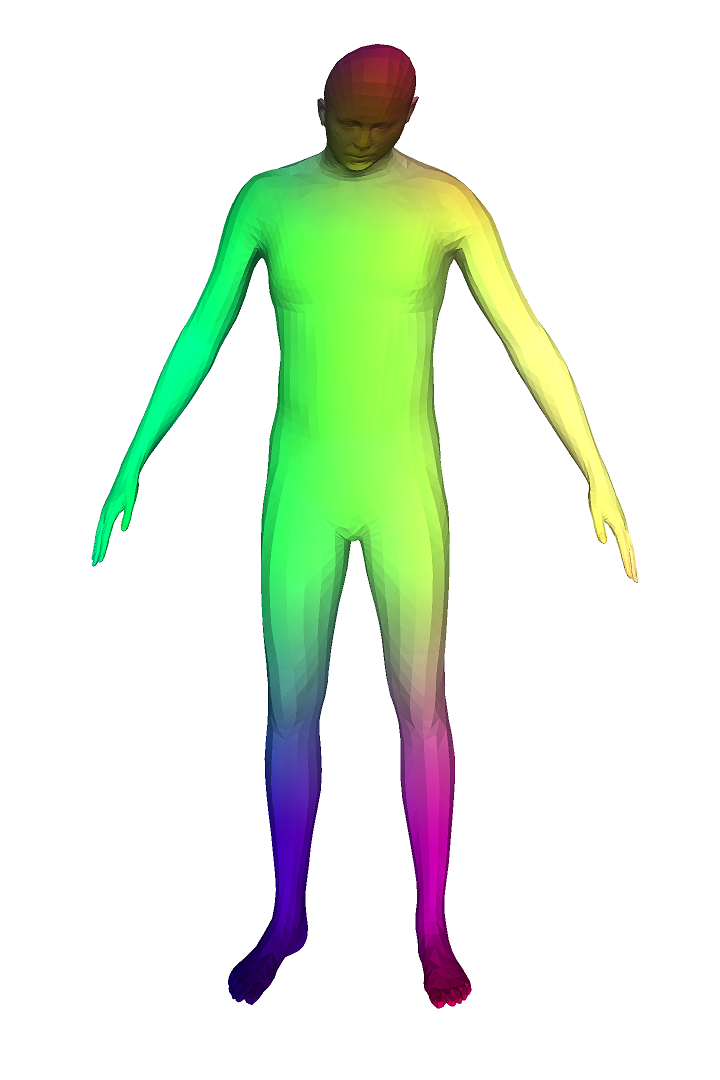}
    \includegraphics[height=3.0cm]{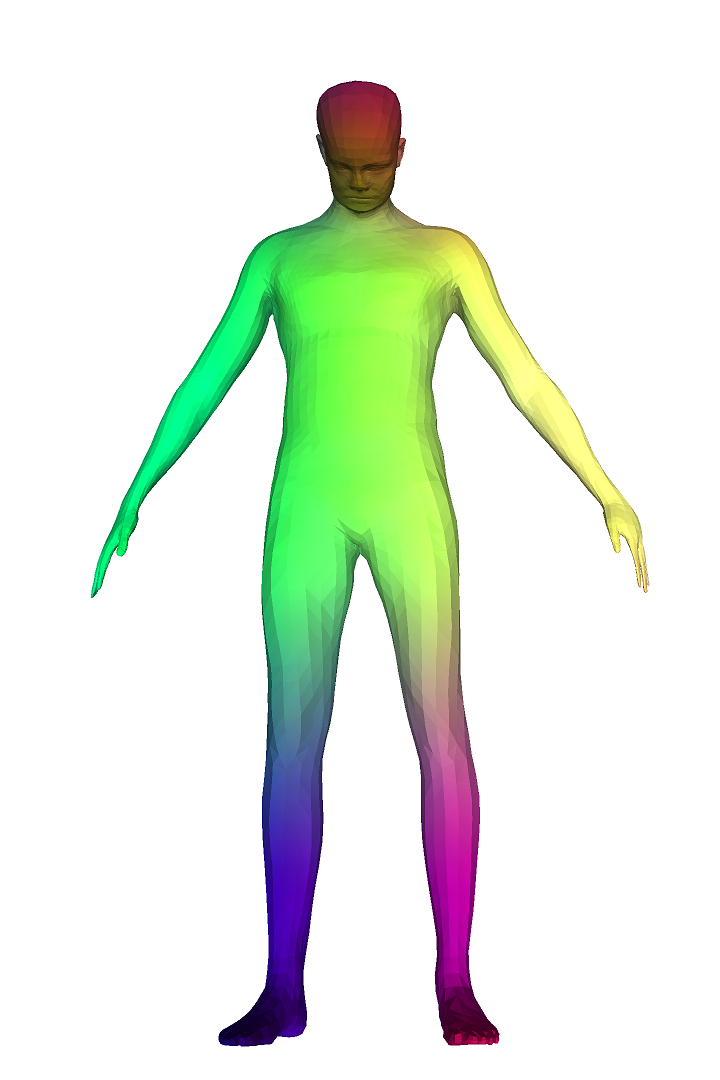}
    
    \includegraphics[height=3.2cm]{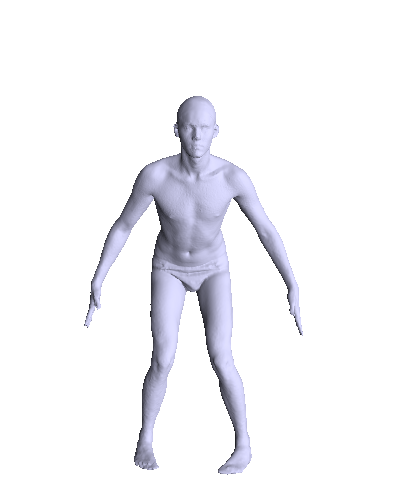}
    \includegraphics[height=3.0cm]{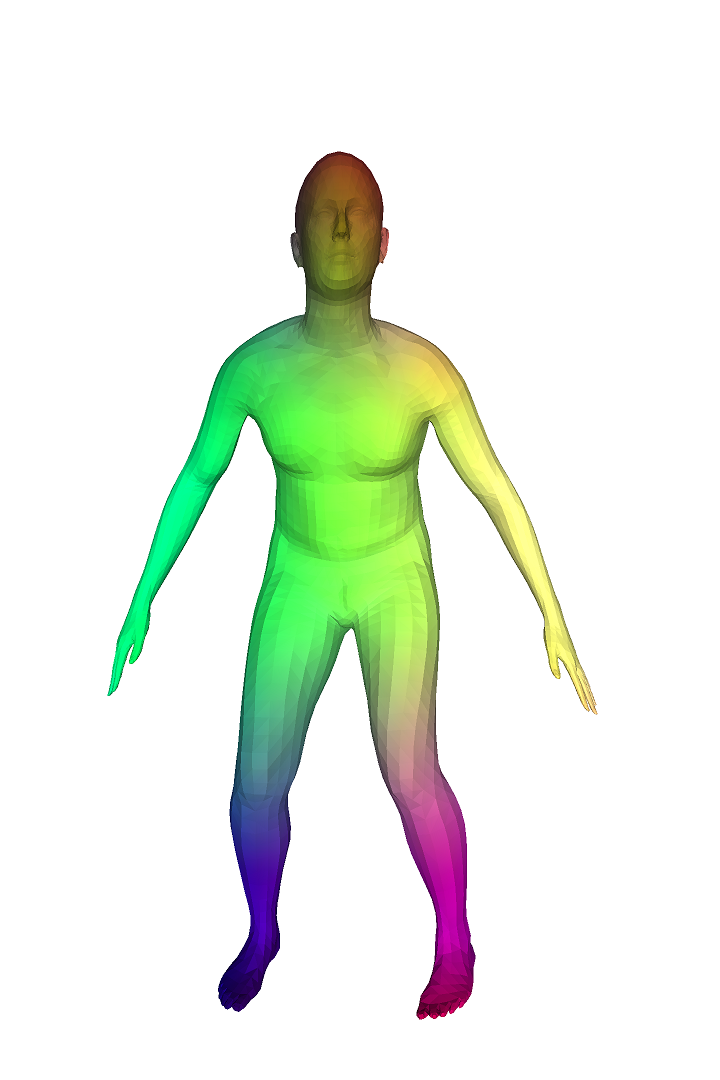}
    \includegraphics[height=3.0cm]{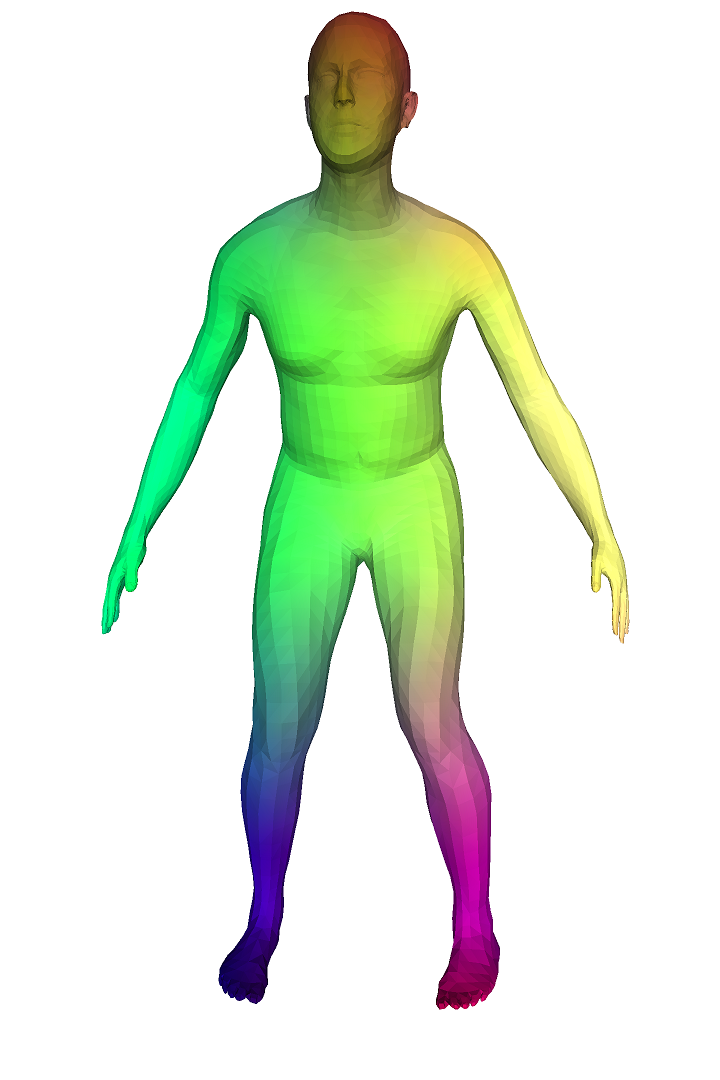}
    \caption{Comparison of Meta Deformation Network and 3D-CODED with Learned templates \cite{deprelle2019learning_elementary} in unoptimized deformation quality. Left: Query Shape, middle: Meta Deformation Networks, right: 3D-CODED with Learned Template]}
    \label{fig:qualitative_comparison_1}
\end{figure}

We show the deformation of the template into the query shapes before the optimization of $\mathbf{E}$, and compare Meta Deformation Network against the non-meta deformation network developed by \index{Deprelle, Theo} T. Deprelle et al. in \cite{deprelle2019learning_elementary}. The figures are included in fig. \ref{fig:qualitative_comparison_1}.

Both approaches give a reasonable unoptimized deformation of a template into the query shape, though both suffer from varying amounts of distortions and imprecision from the actual query shape. The optimization step depicted in \ref{sec:embedding_opt} over $\mathbf{E}$ improves the quality of the deformation.

\subsection{Feature Embedding Optimization}
\label{sec:embedding_opt}
\begin{figure}[ht]
    \centering
    \includegraphics[width=2.2cm]{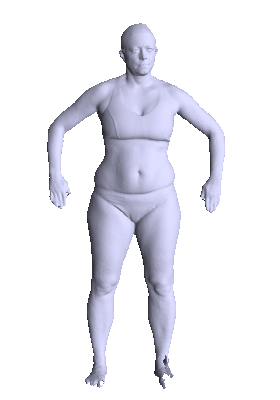}
    \includegraphics[width=2cm]{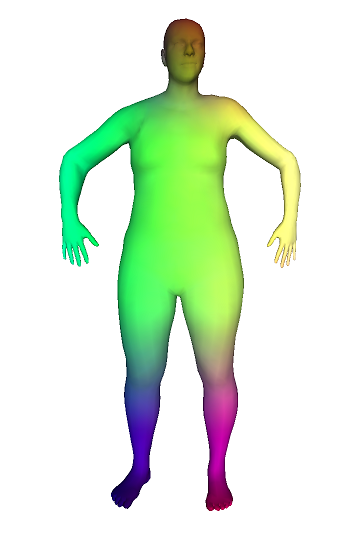}
    \includegraphics[width=2cm]{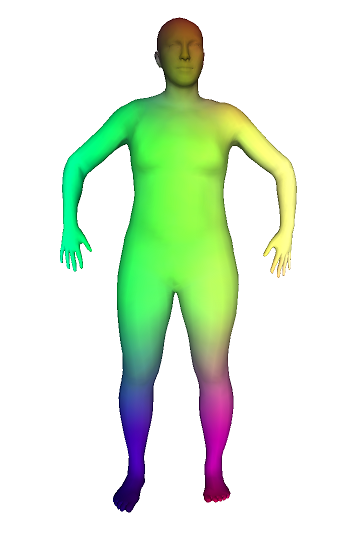}
    \includegraphics[width=2cm]{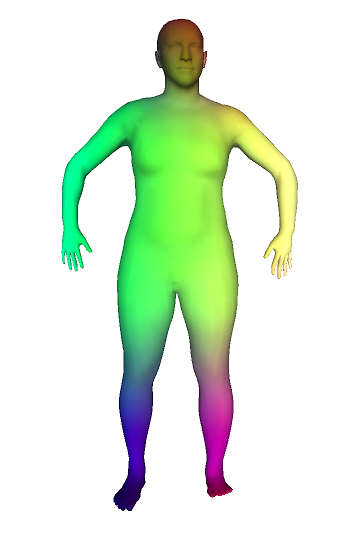}
    \includegraphics[width=2cm]{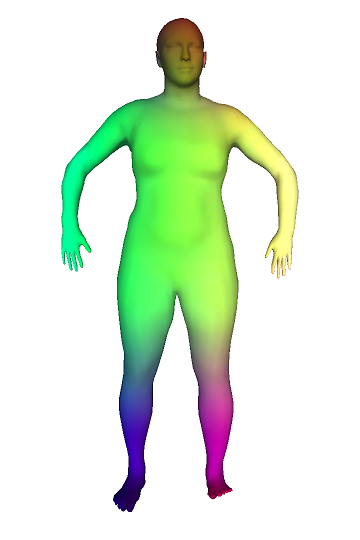}
    \caption{The deformation of the template into the query shape after optimizing $\featureEmbedding$ for different numbers of iterations.
    From left to right: ground-truth query shape, unoptimized deformation, deformation optimized for 200 iterations, 1,000 iterations, and 3,000 iterations. Note the little difference in quality between deformations optimized for 1,000 and 3,000 iterations }
    \label{fig:optimization_iterations}
\end{figure}

We also test how sensitive the quality of the template deformation is to different numbers of optimization steps for $\featureEmbedding$. We qualitatively demonstrate that the deformation continues to improve in quality as $\featureEmbedding$ is optimized for more iterations, but the marginal gain diminishes as the number of iterations increases, with the difference between 1,000 and 3,000 iterations hardly noticeable (See fig. \ref{fig:optimization_iterations}).

The imprecision of the unoptimized deformation suggests that our ShapeNet encoder $f$ is unable to produce an optimal feature embedding $\featureEmbedding$ that fully captures the characteristics of the query shape; the unoptimized deformation has a more generic structure lacking details of the person's body shape, which the post-optimization deformation is able to capture. The relatively weak encoding power of our simplified ShapeNet encoder, can be alleviated by optimizing over $\featureEmbedding$ given $f(\shapeQuery)$ as its initialization. Also note that although the query shape has incomplete data around the person's feet, our decoder has learned to reconstruct the missing parts from the input. 

\subsection{Quantitative Assessment}
\begin{table}[ht]
    \centering
    \begin{tabular}{c|c}
        
        Method & FAUST-Inter Mean Error (cm) \\
        \hline
        Deep functional maps \cite{functional_maps} & 4.82 \\
        Stitched Puppet \cite{Zuffi_2015_CVPR} & 3.12 \\
        3D-CODED w/. Learned 3D Translation Template \cite{deprelle2019learning_elementary} & 3.05 \\
        3D-CODED w/. Learned 3D Deformed Template \cite{deprelle2019learning_elementary} & 2.76 \\
        \hline
        Meta Deformation Network (Ours) & 2.97 \\
    \end{tabular}
    \caption{Comparison of Meta Deformation Network to Network Architectures with Conventional Methods}
    \label{tab:faust_error_comparison}
\end{table}
Though the Meta Deformation Network did not surpass the state-of-the-art performance, it showed an improvement over 3D-CODED w/. Learned 3D Translation Template, an approach that is comparable to ours in that it shares the same simplified PointNet encoder design and also applies a learnable translation matrix to the points of the base template, with the difference being that it its LVC decoder has fixed parameters. This shows quantitatively that having the meta decoder produces more accurate correspondences on FAUST-Inter than does a static-parameter decoder (while also having speedier execution).

We report results from training under a low-resolution template and hot-swapping in a different high-resolution template in inferencing (details for this procedure are given in \ref{sec:transference_high_res}). See the accuracy numbers are shown in table \ref{tab:faust_error_comparison}. 

\subsection{Transference onto Unseen Templates}
\label{sec:transference_high_res}
\begin{table}
    \centering
    \begin{tabular}{c|c}
        
        Template used in inferencing & FAUST-Iter Mean Error (cm) \\
        \hline
        Learned Template (6890 Vertices) & 	3.07 \\
        Unseen Template (176K Vertices) & 	2.97 \\
    \end{tabular}
    \caption{Comparison of infernencing with a low-resolution learned template and a high-resolution new template.}
    \label{tab:faust_hi_res_template}
\end{table}
\begin{figure}
    \centering
    \includegraphics[width=1.8cm]{figures/iterations/test_scan_021.png}
    \includegraphics[width=1.8cm]{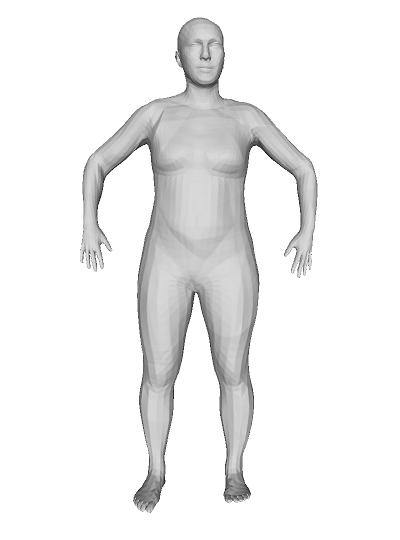}
    \includegraphics[width=1.8cm]{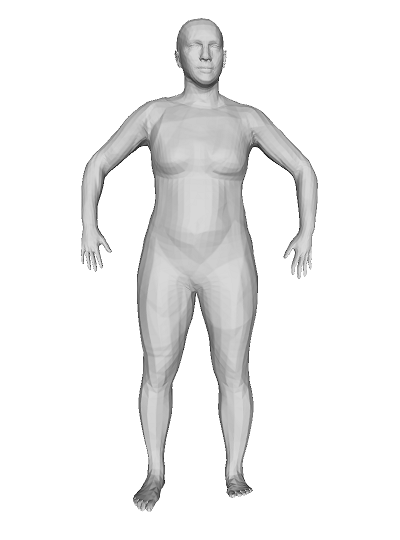}
    \includegraphics[width=1.8cm]{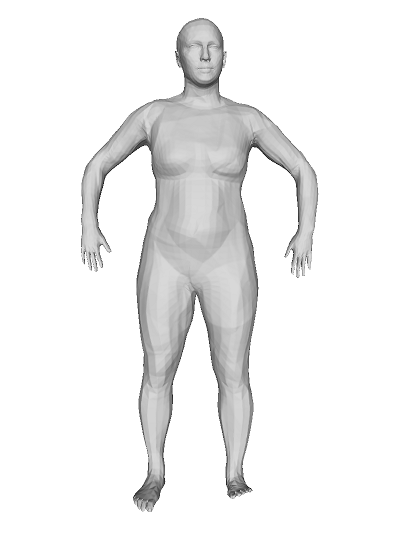}
    \\
    \centering
    \includegraphics[width=1.9cm]{figures/iterations/test_scan_021.png}
    \includegraphics[width=1.8cm]{figures/iterations/reconstruction_021_iter200.png}
    \includegraphics[width=1.8cm]{figures/iterations/reconstruction_021_iter1000.png}
    \includegraphics[width=1.8cm]{figures/iterations/reconstruction_021_iter3000.png}
    \caption{Comparison of decoding from the high-resolution unseen template (top) and learned template (bottom) at 200, 1000, 3000 iterations. The image on the left of both rows is the query shape.}
    \label{fig:unseen_template}
\end{figure}
In the training step, we have trained a feature encoder $f$ and a parameter mapper $h$ to enable $g$ to predict deformations from a known template into unseen query shapes. Here, we also test how well the model performs when given a template and query shapes that are both unseen in training. We run an experimental setup where we find the rotation matrices and the feature embedding initialization for $\shapeQuery$ using the learned template $\shapeTemp$ but swaps in a different, higher-resolution template $\shapeTemp'$ for the optimization step and subsequent correspondence calculation. The learned template $\shapeTemp$ contains 6,890 vertices while the unseen template $\shapeTemp'$ contains slightly over 176,000 vertices. A visualization is given in fig. \ref{fig:unseen_template}. The visual results is consistent with table \ref{tab:faust_hi_res_template}, with little difference in error rate between using a learned template and a high-resolution unseen template.

Results from table \ref{tab:faust_hi_res_template} suggest that $h$ is able to transfer the learned predictions for $\thetaG$ when $g$ operates on a different template. As a result, the meta decoder adapts wells operating on a template that is different from the one used in training and takes advantage of the higher resolution of the new template to produce more accurate correspondences. The high adaptability of the Meta Decoder Network suggests the possibility of variations of the model that use a dynamic template in addition to the dynamic decoder, which we desire to research in future works.

\subsection{Training and Testing Speeds}
\begin{table}[]
    \centering
    \begin{tabular}{c|c}
        
        Method & Time per training iteration(s) \\
        \hline
        3D-CODED w/. Learned 3D Translation Template \cite{groueix20183dcoded} & 0.1964 \\
        Meta Deformation Network (Ours) & 0.1259 \\
    \end{tabular}
    \caption{Comparison of training speeds. Experiment ran with a GTX 1080 Ti and 6-core Intel CPU. Batch size was set to 32 on both runs.}
    \label{tab:faust_training_speed_comparison}
\end{table}
\begin{table}[]
    \centering
    \begin{tabular}{c|c}
        
        Method & Time per pairwise corr. (s) \\
        \hline
        3D-CODED w/. Learned 3D Translation Template \cite{groueix20183dcoded} & 286.42 \\
        Meta Deformation Network (Ours) & 273.66 \\
    \end{tabular}
    \caption{Comparison of testing speeds. Experiment ran with a GTX 1080 Ti and 6-core Intel CPU. Batch size was set to 1 on both runs.}
    \label{tab:faust_testing_speed_comparison}
\end{table}

We compare the training and testing speeds of the Meta Deformation Network against those of 3D-CODED, which has the same encoder structure but uses an LVC decoder. We train both models on the same Extended SURREAL dataset created by \cite{groueix20183dcoded}. 

We find that the Meta Deformation Network took $35.8\%$ less time to carry out an iteration of training (table \ref{tab:faust_training_speed_comparison}) with a batch size set of 32, yet it yields improved accuracy (table \ref{tab:faust_error_comparison}) on the FAUST Inter challenge than did the model from \cite{deprelle2019learning_elementary} that also had a translation-based learnable template.

For testing, we evaluate the time a model takes to compute the correspondence for one pair of 3D scans from FAUST Inter. In testing, our model took $4.45\%$ less time to finish. Here the difference was less dramatic than in training, which we think was due to the under-utilization of the computer's hardware in both methods due to the the latent vector optimization with a batch size of 1.
See table \ref{tab:faust_testing_speed_comparison}.

\section{Conclusion}
Meta Deformation Network is the first implementation of a meta decoder used in a 3-D to 3-D task aim at solving correspondences. It has an encoder-meta-decoder design in which all parameters of the decoder are inferred from a given query shape during both training and testing to improve the deformation of the template. The meta decoder enjoys more adaptability when deforming the template into query shapes compared to the LVC decoder \cite{deprelle2019learning_elementary} with the same encoder. As suggested by testing the Meta Deformation Network on the FAUST-Inter challenge, the meta decoder achieves better correspondence accuracy with the added benefit of speedier training and inferencing. 

\clearpage
\bibliographystyle{splncs04}
\bibliography{main}

\begin{thebibliography}{10}
\providecommand{\url}[1]{\texttt{#1}}
\providecommand{\urlprefix}{URL }
\providecommand{\doi}[1]{https://doi.org/#1}

\bibitem{bertinetto2016learning}
Bertinetto, L., Henriques, J.F., Valmadre, J., Torr, P.H.S., Vedaldi, A.:
  Learning feed-forward one-shot learners (2016)

\bibitem{Bogo:CVPR:2014}
Bogo, F., Romero, J., Loper, M., Black, M.J.: {FAUST}: Dataset and evaluation
  for {3D} mesh registration. In: Proceedings IEEE Conf. on Computer Vision and
  Pattern Recognition (CVPR). IEEE, Piscataway, NJ, USA (Jun 2014)

\bibitem{brab2016dynamic}
Brabandere, B.D., Jia, X., Tuytelaars, T., Gool, L.V.: Dynamic filter networks
  (2016)

\bibitem{chang2015shapenet}
Chang, A.X., Funkhouser, T., Guibas, L., Hanrahan, P., Huang, Q., Li, Z.,
  Savarese, S., Savva, M., Song, S., Su, H., Xiao, J., Yi, L., Yu, F.:
  Shapenet: An information-rich 3d model repository (2015)

\bibitem{deprelle2019learning_elementary}
Deprelle, T., Groueix, T., Fisher, M., Kim, V.G., Russell, B.C., Aubry, M.:
  Learning elementary structures for 3d shape generation and matching (2019)

\bibitem{fan2016point}
Fan, H., Su, H., Guibas, L.: A point set generation network for 3d object
  reconstruction from a single image (2016)

\bibitem{groueix20183dcoded}
Groueix, T., Fisher, M., Kim, V.G., Russell, B.C., Aubry, M.: 3d-coded : 3d
  correspondences by deep deformation (2018)

\bibitem{groueix2018atlasnet}
Groueix, T., Fisher, M., Kim, V.G., Russell, B.C., Aubry, M.: Atlasnet: A
  papier-mâché approach to learning 3d surface generation (2018)

\bibitem{7299117_dynamic_weather_prediction}
{Klein}, B., {Wolf}, L., {Afek}, Y.: A dynamic convolutional layer for short
  rangeweather prediction. In: 2015 IEEE Conference on Computer Vision and
  Pattern Recognition (CVPR). pp. 4840--4848 (June 2015).
  \doi{10.1109/CVPR.2015.7299117}

\bibitem{littwin2019deep}
Littwin, G., Wolf, L.: Deep meta functionals for shape representation (2019)

\bibitem{SMPL:2015}
Loper, M., Mahmood, N., Romero, J., Pons-Moll, G., Black, M.J.: {SMPL}: A
  skinned multi-person linear model. ACM Trans. Graphics (Proc. SIGGRAPH Asia)
  \textbf{34}(6),  248:1--248:16 (Oct 2015)

\bibitem{mitchell2019higherorder}
Mitchell, E., Engin, S., Isler, V., Lee, D.D.: Higher-order function networks
  for learning composable 3d object representations (2019)

\bibitem{functional_maps}
Ovsjanikov, M., Ben-Chen, M., Solomon, J., Butscher, A., Guibas, L.: Functional
  maps: A flexible representation of maps between shapes. ACM Transactions on
  Graphics - TOG  \textbf{31} (07 2012). \doi{10.1145/2185520.2185526}

\bibitem{park2019deepsdf}
Park, J.J., Florence, P., Straub, J., Newcombe, R., Lovegrove, S.: Deepsdf:
  Learning continuous signed distance functions for shape representation (2019)

\bibitem{qi2016pointnet}
Qi, C.R., Su, H., Mo, K., Guibas, L.J.: Pointnet: Deep learning on point sets
  for 3d classification and segmentation (2016)

\bibitem{richter2018matryoshka}
Richter, S.R., Roth, S.: Matryoshka networks: Predicting 3d geometry via nested
  shape layers (2018)

\bibitem{7410424_conditioned_super_res}
{Riegler}, G., {Schulter}, S., {Rüther}, M., {Bischof}, H.: Conditioned
  regression models for non-blind single image super-resolution. In: 2015 IEEE
  International Conference on Computer Vision (ICCV). pp. 522--530 (Dec 2015).
  \doi{10.1109/ICCV.2015.67}

\bibitem{varol17_surreal}
Varol, G., Romero, J., Martin, X., Mahmood, N., Black, M.J., Laptev, I.,
  Schmid, C.: Learning from synthetic humans. In: CVPR (2017)

\bibitem{yang2017foldingnet}
Yang, Y., Feng, C., Shen, Y., Tian, D.: Foldingnet: Point cloud auto-encoder
  via deep grid deformation (2017)

\bibitem{zeng2018inferring}
Zeng, W., Karaoglu, S., Gevers, T.: Inferring point clouds from single
  monocular images by depth intermediation (2018)

\bibitem{Zuffi_2015_CVPR}
Zuffi, S., Black, M.J.: The stitched puppet: A graphical model of 3d human
  shape and pose. In: The IEEE Conference on Computer Vision and Pattern
  Recognition (CVPR) (June 2015)

\end{thebibliography}
\end{document}